\documentclass{article}
\usepackage{spconf,amsmath,epsfig}
\usepackage{times}
\usepackage{color}

\usepackage{hyperref}
\hypersetup{
    colorlinks=true,
    linkcolor=blue,
    filecolor=magenta,      
    urlcolor=magenta,
}

\title{Leveraging Category Information for \\Single-Frame Visual Sound Source Separation}
\twoauthors
 {Lingyu Zhu}
	{Computer Vision Group\\
	Tampere University, Finland\\
	lingyu.zhu@tuni.fi}
 {Esa Rahtu}
	{Computer Vision Group\\
	Tampere University, Finland\\
	esa.rahtu@tuni.fi}
\begin{document}
%
\maketitle
\begin{abstract}
Visual sound source separation aims at identifying sound components from a given sound mixture with the presence of visual cues. Prior works have demonstrated impressive results, but with the expense of large multi-stage architectures and complex data representations (e.g. optical flow trajectories). In contrast, we study simple yet efficient models for visual sound separation using only a single video frame. Furthermore, our models are able to exploit the information of the sound source category in the separation process. To this end, we propose two models where we assume that i) the category labels are available at the training time, or ii) we know if the training sample pairs are from the same or different category. The experiments with the $\textit{MUSIC}$ dataset show that our model obtains comparable or better performance compared to several recent baseline methods. The code is available at \href{https://github.com/ly-zhu/Leveraging-Category-Information-for-Single-Frame-Visual-Sound-Source-Separation}{https://github.com/ly-zhu/Leveraging-Category-Information-for-Single-Frame-Visual-Sound-Source-Separation}.
\end{abstract}

%


\section{Introduction}
\label{sec:intro}

Human perceives a scene by looking and listening, which requires different senses to capture multiple modalities (e.g. audio and vision) and the ability of associating the received signals. Likewise, recent machine learning methods~\cite{Afouras20b,morgado2020learning,gan2020look,ephrat2018looking,zhao2018sound,arandjelovic2017look} utilise multi-modal data to address the complex perception tasks. In this work, we are interested in the task of self-supervised audio-visual sound separation, where the objective is to distinguish original sound components via joint audio-appearance learning. Moreover, we study how the sound sources could be localized from the associated visual data. 

Recent works~\cite{Afouras20b,zhu2020visually,gan2020music,zhao2019sound,xu2019recursive,gao2018learning,ephrat2018looking,owens2018audio,zhao2018sound} achieve remarkable results on various sound separation and localization tasks by conditioning on the associated appearance and motion cues. While the motions may be crucial for visual sound separation under certain circumstances, the single frame based approaches perform surprisingly well as demonstrated in \cite{zhao2018sound,zhu2020visually}. 

\begin{figure}[tp]
    \centering
    \includegraphics[width=1.0\linewidth]{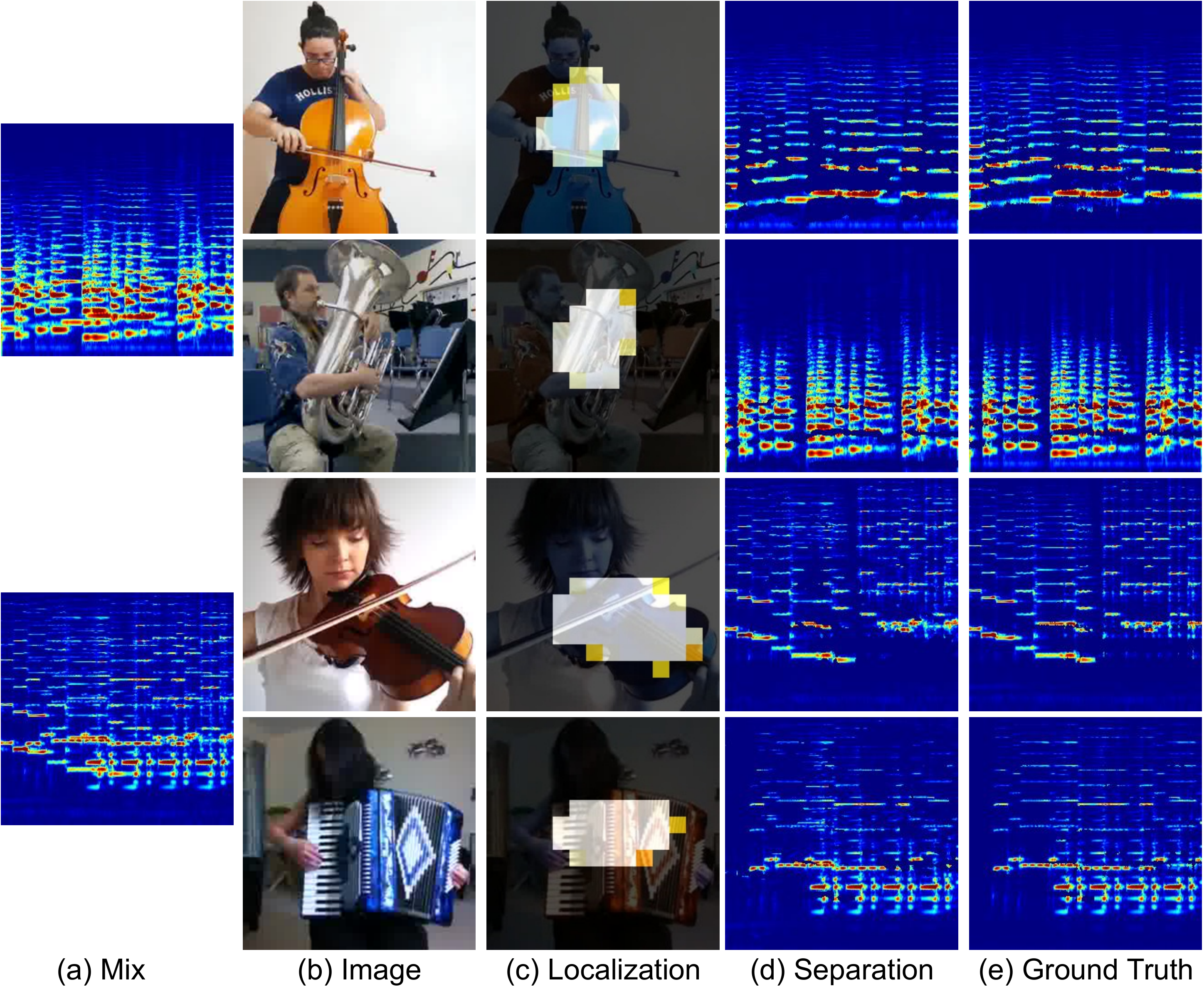}
    \caption{Examples of sound source separation (d) and localization (c) conditioning on a single video frame.}
    \label{fig:locsep_vis}
\end{figure}

We hypothesise that the performance is mostly obtained by inferring the visual category of the sound source (e.g. the instrument type) and then extracting the category specific mixture components. We are interested in how far the performance of the single-frame based source separation models (see e.g. Fig.~\ref{fig:locsep_vis}) can be pushed and how the categorical information can be optimally utilised in this framework. 

\begin{figure*}[tp]
    \centering
    \includegraphics[width=0.90\linewidth]{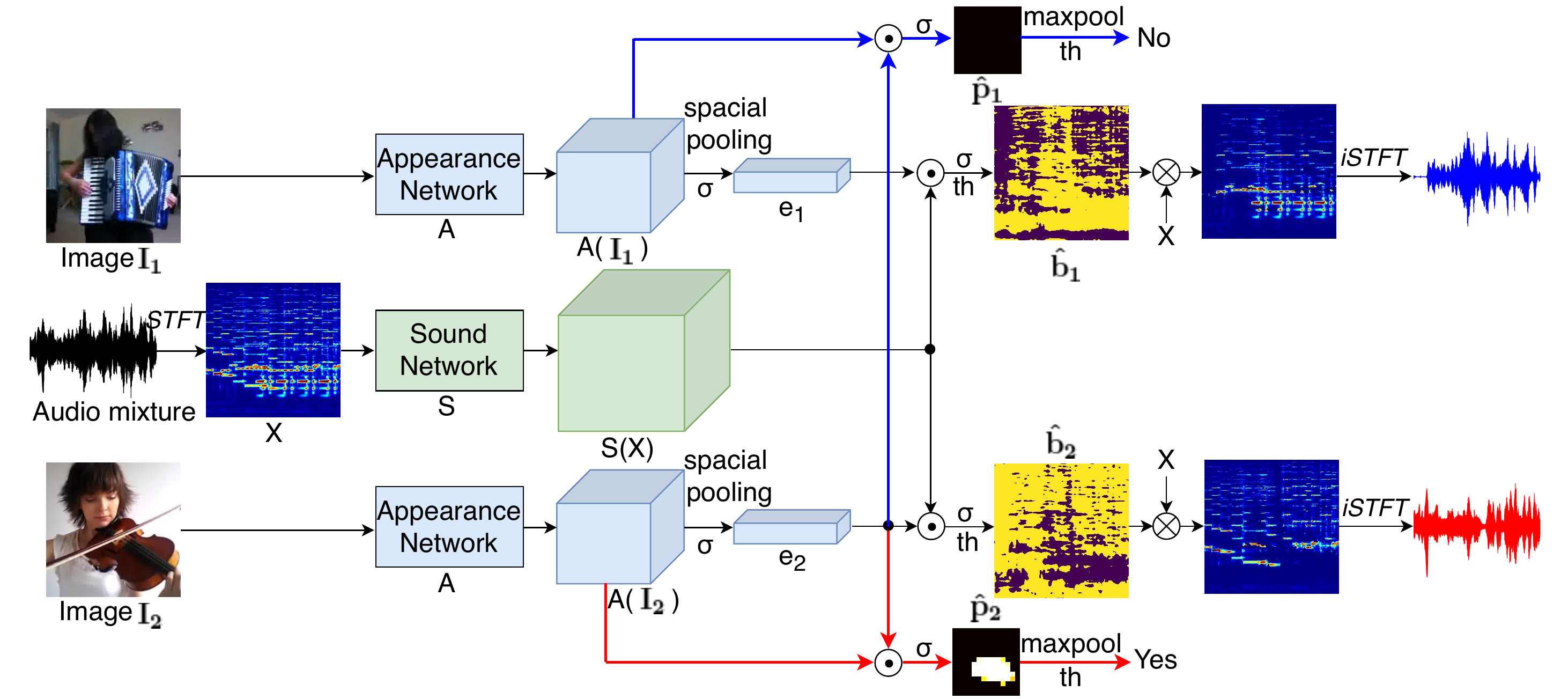}
    \caption{The framework of single frame visual sound source separation and localization system. The appearance network $A$ converts the input image $I$ (a random frame of a sequence video) to visual feature maps $A(I)$ and further, with a spacial pooling, to a compact representation $e$. The sound network $S$ splits the mixture spectrogram $X$ into a set of feature maps $S(X)$. A linear combination of appearance representation $e$ and sound features maps $S(X)$ produces a sound separation mask $\hat{b}$. The appearance attention module (red and blue arrows) is formed by a scalar product between the appearance representation $e$ and appearance feature maps $A(I)$. The appearance attention module produces a source location mask $\hat{p}$.}
    \label{fig:locsep}
\end{figure*}

In this paper, we study a simple yet effective single frame visual sound separation approach, which is able to exploit the categorical information of the sources in multiple ways. Given two different sources, we first separate sounds using the appearance embedding of each source learned using the appearance network. Then, we introduce a light yet efficient appearance attention module (Fig.~\ref{fig:locsep}) that enhances the semantic distinction of the predicted appearance embedding by predicting the correspondence between the appearance embedding and the appearance feature maps. We show that the appearance attention module can greatly improve the sound source separation performance compared to the baseline systems. In addition, the proposed appearance attention module can precisely locate the sound locations (see e.g. Fig.~\ref{fig:locsep_vis}) without additional computations. Finally, we assess to what extend the categorical information could aid the source separation task. To this end, we utilise the category labels for training the appearance embedding modules. Moreover, we examine the extreme case, where the category labels of the test samples would be known.

\section{Related Work}
\label{sec:related_work}

\subsection{Audio-Visual Sound Source Separation} 

Researchers have recently proposed various learning-based approaches to include the visual signal to the task of sound separation. Ephrat~\textit{et al.}~\cite{ephrat2018looking} extracted face embeddings using a pre-trained face recognition model to facilitate speech separation. Similarly, Gao~\textit{et al.}~\cite{gao2019co} trained an object detector to localize objects in all video frames to improve the sound separation quality. Zhao~\textit{et al.}~\cite{zhao2018sound,zhao2019sound} learned to separate sounds with appearance and motions. Xu~\textit{et al.}~\cite{xu2019recursive} separated sounds by recursively removing the sounds with large energy from sounds mixture. Gan~\textit{et al.}~\cite{gan2020music} associated body and finger movements with audio signals by learning a keypoint-based structured representation from a Graph CNN. Zhu~\textit{et al.}~\cite{zhu2020visually} proposed a cascaded opponent filter to utilize visual features of all sources to look for incorrectly assigned sound components from opponent sources.

\subsection{Sound Source Localization}
Localizing sound sources entails identifying the regions where the sound comes from. Effort had been put to explore the audio-visual synchrony~\cite{hershey2000audio} and canonical correlations~\cite{kidron2005pixels}. Most recently, Arandjelovic~\textit{et al.}~\cite{arandjelovic2018objects} visualized sound location by computing the similarity between the audio and all visual embeddings. \cite{owens2018learning,owens2018audio} applied the class activation map for localizing ambient sounds. ~\cite{zhao2018sound,zhao2019sound,xu2019recursive} visualized sound sources by calculating the sound volume at each spatial location. Gao~\textit{et al.}~\cite{gao2019co} localized potential source regions via a separate object detector. Zhu~\textit{et al.}~\cite{zhu2020visually} located sound sources by learning to identify a minimum set of input pixels to produce almost identical output as for the entire image. 

\section{Approach}
\label{sec:method}

This section describes our single frame visual sound source separation system. The model is illustrated in Fig.~\ref{fig:locsep} and it consists of four main components: the appearance network, sound network, sound source separation module, and the appearance attention module. The appearance and sound networks encode the visual and audio inputs into feature embeddings, respectively. The embeddings are subsequently fused to obtain a binary mask for each source indicating which parts of the mixture spectrogram belong to the corresponding source. The overall architecture is inspired by~\cite{zhao2018sound}, but we make important changes to the component structures and introduce a new appearance attention module to incorporate categorical information. The following subsections describe further details of each component and the learning objective.

\subsection{Appearance Network}
\label{sec:app3}
The visual cues are extracted from a single frame $I$ (randomly sampled from the input video). We use Res-18 and Res-50~\cite{he2016deep} architectures as two alternatives for the appearance network. The appearance network $A$ converts the input image $I$ of size $3 \times H \times W$ to feature maps $A(I)$ of size $K \times H/16 \times W/16$. With a spatial max pooling and sigmoid operation, we get a compact visual representation $e$ of size $1 \times K$.

\subsection{Sound Network}
\label{sec:sound3}
The sound network S is implemented using U-Net~\cite{ronneberger2015u} or MobileNetV2 (MV2)~\cite{sandler2018mobilenetv2} architectures. The input to the sound network is a mixture audio, which is represented as spectrograms that obtained from the audio stream using Short-time Fourier Transform (STFT). The sound network $S$ converts the input spectrogram $X$ of size $1 \times HS \times WS$ into a set of feature maps $S(X)$ of size $K \times HS \times WS$. The K equals to the K of the appearance representation from appearance network.

\subsection{Sound Source Separation}
\label{sec:sep3}
The sound source separation is achieved by a linear combination between the learned appearance representation $e$ and the sound feature maps $S(X)$, as follows
\begin{equation}
    \centering
    \label{eq:1}
        \hat{b} = th(\sigma(\sum_{k} \, e_{k} \odot S(X)_{k}))
\end{equation}
where $e_{k}$ is the $k$-th element of the appearance representation, and $S(X)_{k}$ is the $k$-th sound network feature map. $\odot$ indicates scalar product. $\sigma$ denotes the sigmoid operation. We get the predicted binary mask $\hat{b}$ by setting a threshold of 0.5.

\subsection{Appearance Attention Module}
\label{sec:att3}
The performance of the appearance network is hindered by the appearance similarity and the noise within the video sequences. In order to enhance the distinction of the predicted semantic representations, we introduce an appearance attention branch with an auxiliary loss to the appearance network. The appearance attention module is depicted as red and blue arrows in Fig.~\ref{fig:locsep}. 

The appearance attention module exploits the knowledge of the source types. In particular, we assume to know, at the training time, if the sources $I_{1}$ and $I_{2}$ are from the same or different type (e.g. same or different instrument types). As is shown in Fig.~\ref{fig:locsep}, the appearance attention module is optimized by predicting whether the appearance embedding $e$ and appearance feature maps $A(I)$ are from the same categories (red arrows, e.g. $e_{2}$ and $A(I_{2})$) or not (blue arrows, e.g. $e_{2}$ and $A(I_{1})$). The scalar product between the appearance embedding and appearance feature maps of same category will locate the sound sources (e.g. $\hat{p}_{2}$), and of different categories will return a blank mask (e.g. $\hat{p}_{1}$). The output of the appearance attention module is described as below,

\begin{equation}
\label{eq:2}
  \left.\begin{aligned}
  \hat{p}_{pos} = \sigma(\sum_{k} \, e_{nk} \odot A(I_{n})_{k})\\
  \hat{p}_{neg} = \sigma(\sum_{k} \, e_{nk} \odot A(I_{m})_{k})
\end{aligned}\right\} 
    \begin{aligned}
      n, m \in &[0, N-1], \\
      m &\neq n
    \end{aligned}
\end{equation}
where $\hat{p}$ is the predicted location mask of the appearance attention module. $A(I_{n})$ is the appearance feature maps of n-th video, $e_{n}$ is its corresponding appearance embedding that derived from $A(I_{n})$ by a spatial pooling and sigmoid operation. N is the number of sounds in sound mixture. k $\in$ [0, K-1]. K is the number of elements in appearance representation $e$ as well as the channel number of feature maps $A(I)$.

\subsection{Visual Category Information in Sound Separation}
\label{sec:classifier3}
To further study the application of categorical information in the visual source separation, we train an appearance classifier to predict the source type when given the object category information at the training time. For sound source separation, we replace the appearance representation $e$ in Eq.~\ref{eq:1} (see Sec.~\ref{sec:sep3}) with the appearance embedding that produced by the appearance classifier from a single frame. Finally, we investigate the extreme case, where the category information of the sources would be available at the test time. For this purpose, we replace the appearance embedding $e$ with the one-hot encoding of the category.

\subsection{Learning Objective}

The learning objective of our system is to estimate a binary mask $\hat{b}$ (Eq.~(\ref{eq:1})) to separate the target sound from mixture and to predict a location mask $\hat{p}$ (Eq.~(\ref{eq:2})) for locating the sounding sources. The ground truth mask $b$ of sound separation is calculated based on whether the target sound is the dominant component in the mixture spectrogram $X$. We add an appearance attention module with an auxiliary loss to the appearance network. The ground truth $p$ of the appearance attention module is defined by whether the appearance embedding and the appearance feature maps are from the same categories or not. The model parameters are optimised with respect to the Binary Cross Entropy (BCE) loss. More specifically,

\begin{equation}
    \centering
    \label{eq:3}
        \mathcal{L} = \textit{BCE}(\hat{b}, b) + \textit{BCE}(maxpool(\hat{p}), p)
\end{equation}
where $\hat{p}$ is the predicted location mask of the appearance attention module.

\begin{figure}[!tbp]
    \centering
        \includegraphics[width=1.0\linewidth, keepaspectratio]{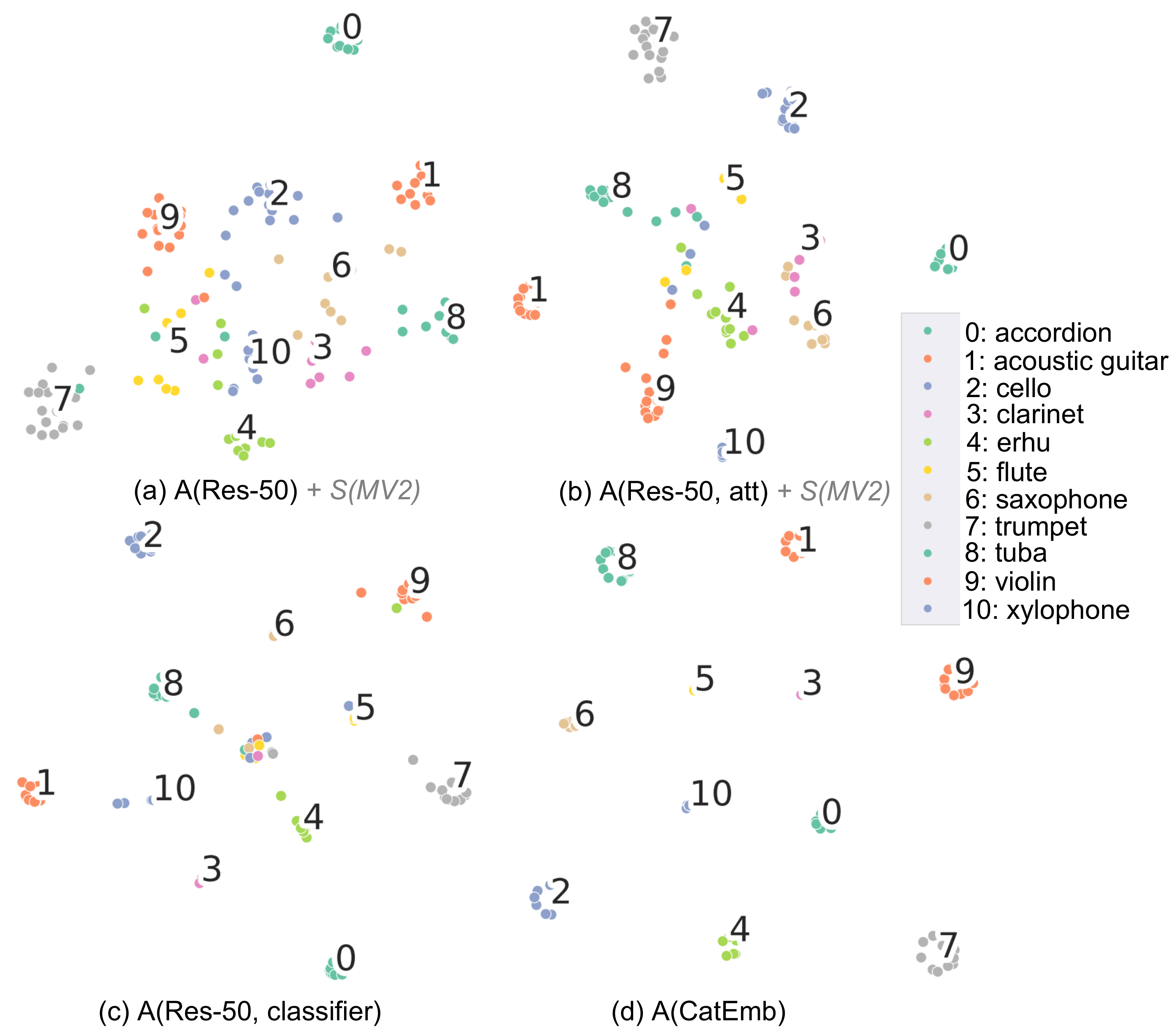}
    \caption{t-SNE visualization of (a) A(Res-50), (b) A(Res-50, att), (c) A(Res-50, classifier), and (d) A(CatEmb).}
\label{fig:tsne}
\end{figure}

\section{Experiments}
\label{sec:exps}

We evaluate the studied methods using the public dataset MUSIC~\cite{zhao2018sound}. The performance of the final sound source separation is measured in terms of standard metrics: Signal to Distortion Ratio (SDR), Signal to Interference Ratio (SIR), and Signal to Artifact Ratio (SAR). Higher value is better for all metrics.

\subsection{Dataset and Implementation details}

MUSIC~\cite{zhao2018sound} dataset is a high quality dataset of musical instruments. The dataset has little off-screen noise and contains 714 untrimmed YouTube videos which span 11 instrumental categories. We follow the same dataset set up as \cite{zhu2020visually}. During training, we randomly select $N$=2 different types of videos with paired frame and audio. The audio mixture $X$ is obtained by adding audio signals from the $N$ different videos. We extract video frames at 8fps and sub-sample each audio signal at 11kHz and randomly crop an audio clip of 6 seconds for training. A Time-Frequency (T-F) spectrogram of size $512\times256$ is obtained by applying Short-time Fourier Transform (STFT), with a Hanning window size of 1022 and a hop length of 256, to the input sound clip. We further re-sample this spectrogram to a T-F representation of size $256\times256$ on a log-frequency scale. The final separated sound is achieved by adding an inverse Short-time Fourier Transform (iSTFT) to the predicted component spectrogram.

\setlength{\tabcolsep}{4pt}
\begin{table}[!tbp]
    \centering
    \begin{tabular}{llll}
        \hline\noalign{\smallskip}
        Models & SDR & SIR & SAR\\
        \noalign{\smallskip}
        \hline\hline
        \noalign{\smallskip}
            \textit{A(Res-18) + S(U-Net)} & \textit{5.38} & \textit{11.00} & \textit{9.77}\\
            A(Res-18, att) + S(U-Net) & 6.48 & 12.06 & 10.31\\
            A(Res-18, classifier) + S(U-Net) & 7.13 & 13.74 & 10.14\\
            \hline
            \textit{A(Res-50) + S(U-Net)} & \textit{5.88} & \textit{11.09} & \textit{10.73}\\
            A(Res-50, att) + S(U-Net) & 7.14 & 12.83 & 10.93\\
            A(Res-50, classifier) + S(U-Net) & 8.38 & 14.94 & 10.85\\
            \hline
            A(CatEmb) + S(U-Net) & 8.55 & 14.98 & 11.21 \\
            \hline \hline
            \textit{A(Res-18) + S(MV2)} & \textit{7.73} & \textit{13.48} & \textit{11.55}\\
            A(Res-18, att) + S(MV2) & 9.22 & 15.22 & 12.62\\
            A(Res-18, classifier) + S(MV2) & 10.06 & 16.82 & 12.66\\
            \hline
            \textit{A(Res-50) + S(MV2)} & \textit{7.95} & \textit{13.66} & \textit{12.16}\\
            A(Res-50, att) + S(MV2) & 9.41 & 15.56 & 12.66\\
            A(Res-50, classifier) + S(MV2) & 10.59 & 17.23 & 12.75\\
            \hline
            A(CatEmb) + S(MV2)& \textbf{10.74} & \textbf{17.29} & \textbf{13.04} \\
        \hline
    \end{tabular}
    \caption{The sound source separation results on MUSIC test set. A: appearance network, S: sound network, and att: appearance attention module. The best results are bolded.}
    \label{table:sound}
\end{table}
\setlength{\tabcolsep}{1.4pt}

\subsection{Sound Source Separation with Appearance Network}

We combine the appearance network $A$ of Res-18 and Res-50 with sound network $S$ of U-Net and MV2 as four models\footnote{without the red and blue arrows of Fig.~\ref{fig:locsep}} to compare against: A(Res-18) + S(U-Net), A(Res-18) + S(MV2), A(Res-50) + S(U-Net), and A(Res-50) + S(MV2). We report the corresponding sound separation metrics in Table~\ref{table:sound} (\textit{italic}).

\subsection{Sound Source Separation with Appearance Attention Module}
\label{sec:att}

As is shown in Table~\ref{table:sound}, with the same appearance network, the higher capacity the sound network has, the better performance the system achieves, e.g. moving from A(Res-18) + S(U-Net) to A(Res-18) + S(MV2) results in SDR: 2.35dB performance improvement. However, with the same sound network, having the appearance network of higher capacity results in similar performance improvement, e.g. the improvement from A(Res-18) + S(MV2) to A(Res-50) + S(MV2) is only SDR: 0.22dB. Thus, we hypothesize that the appearance embedding that predicted from the appearance network could be further improved for sound separation. 

To study this, we introduced a light yet efficient appearance attention module to emphasize the semantic distinction of the learned appearance embeddings. We assess the performance of the appearance attention module (denoted as $att$) for the sound separation task (sound source localization in Sec~\ref{sec:loc_vis}) in Table~\ref{table:sound}. The improvement, e.g. SDR: 1.49dB, of A(Res-18, att) + S(MV2) compared to its counterpart A(Res-18) + S(MV2), indicates a clear advantage from the appearance attention module.

\setlength{\tabcolsep}{4pt}
\begin{table}[t]
    \centering
    \begin{tabular}{llll}
        \hline\noalign{\smallskip}
        Models & SDR & SIR & SAR\\
        \noalign{\smallskip}
        \hline\hline
        \noalign{\smallskip}
            SoP~\cite{zhao2018sound} & 5.38 & 11.00 & 9.77\\
            SoM~\cite{zhao2019sound} & 4.83 & 11.04 & 8.67\\
            MP-Net~\cite{xu2019recursive} & 5.71 & 11.36 & 10.45\\
            Co-Separation~\cite{gao2019co} & 7.38 & 13.7 & 10.8\\
            COF~\cite{zhu2020visually} & 10.07 & 16.69 & \textbf{13.02}\\
            \hline
            A(Res-18) + S(MV2) & 7.73 & 13.48 & 11.55\\
            A(Res-18, att) + S(MV2) & 9.22 & 15.22 & 12.62\\
            A(Res-18, classifier) + S(MV2) & 10.06 & 16.82 & 12.66\\
            \hline
            A(Res-50) + S(MV2) & 7.95 & 13.66 & 12.16\\
            A(Res-50, att) + S(MV2) & 9.41 & 15.56 & 12.66\\
            A(Res-50, classifier) + S(MV2) & \textbf{10.59} & \textbf{17.23} & 12.75\\
            A(CatEmb) + S(MV2)& \textbf{10.74} & \textbf{17.29} & \textbf{13.04} \\
        \hline
    \end{tabular}
    \caption{Single frame visual sound source separation results in comparison with recent approaches SoP~\cite{zhao2018sound}, SoM~\cite{zhao2019sound}, MP-Net~\cite{xu2019recursive}, Co-Separation~\cite{gao2019co}, and COF~\cite{zhu2020visually}. A: appearance network, S: sound network, and att: appearance attention module. The top 2 results are bolded.}
    \label{table:comparison}
\end{table}
\setlength{\tabcolsep}{1.4pt}

\subsection{Sound Source Separation using Category Embeddings}
\label{sec:binary}

In this section, we investigate how the category information alone could aid for the sound source separation. For this purpose, we encode the known source category information of a visual frame into binary embedding, namely, Category Embedding (CatEmb). The CatEmb will be a replacement of the learned appearance embedding $e$ from appearance network. At the training phase, we adopt the CatEmb as the appearance cues to separate the target sound components from the sound mixture with the sound network. As is shown in Table~\ref{table:sound}, with the CatEmb, the sound network of U-Net~\cite{ronneberger2015u} and MV2~\cite{sandler2018mobilenetv2} attain the performance of 8.55dB and 10.74dB in SDR respectively. The results suggest how much the existing appearance network could be further improved.

\subsection{Sound Source Separation with Appearance Classifier}
\label{sec:classifier}

How far could we push the appearance network embedding towards the CatEmb? To answer this question, we first train an appearance classifier when providing the category information, and then adopt it for the sound separation task. Its quantitative result is reported in Table~\ref{table:sound}. With the appearance classifier, the system pushes its sound separation performance further towards the models that use CatEmb, e.g. the scores of SDR: 10.59 of A(Res-50, classifier) + S(MV2) in comparison with the SDR: 10.74 of A(CatEmb) + S(MV2). We take the framework of A(Res-50) + S(MV2) as an example to visualize the appearance embedding on different conditions (e.g. appearance network, appearance attention, appearance classifier, and CatEmb) with t-SNE~\cite{maaten2008visualizing} in Fig.~\ref{fig:tsne}. As we can see, the compactness of both the intra- and inter-class of A(Res-50) embedding is limited. From the A(Res-50) to A(Res-50, att), and A(Res-50, classifier), the learned appearance embedding is pushed more close to the CatEmb in Fig.~\ref{fig:tsne}(d).

\begin{figure}[!tbp]
    \centering
        \includegraphics[width=1.0\linewidth,keepaspectratio]{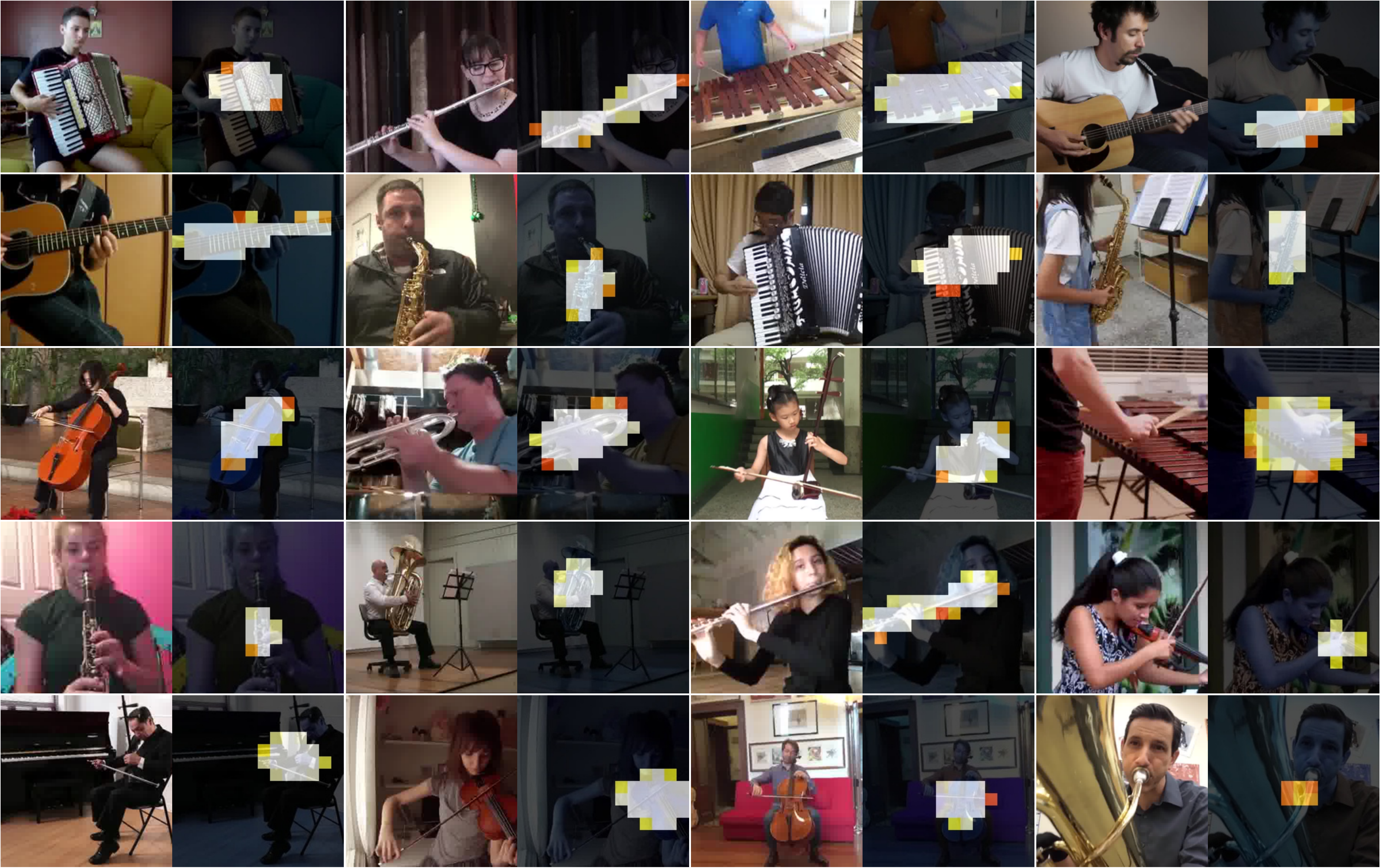}
    \caption{Visualizing sound source locations by the appearance attention module.}
\label{fig:loc_vis}
\end{figure}

\subsection{Sound Source Localization with Appearance Attention Module}
\label{sec:loc_vis}

Given a sound mixture and a single frame, we use the spacial-pooled appearance representation to give self attention to the appearance features (appearance attention module) for localizing the sounding objects. It is shown as red arrows in Fig.~\ref{fig:locsep}. We visualize the sound source location examples in Fig.~\ref{fig:loc_vis} by applying the appearance attention module, which precisely localizes sound sources. We display the spatial location in heatmaps on the input image.

\subsection{Comparison with State-of-the-Art}

We compare our single frame visual sound source separation models with SoP~\cite{zhao2018sound}, SoM~\cite{zhao2019sound}, MP-Net~\cite{xu2019recursive}, Co-Separation~\cite{gao2019co}, and COF~\cite{zhu2020visually} using MUSIC dataset in Table~\ref{table:comparison}. Note that the compared approaches either utilize complex data representations (e.g. optical flow trajectories) or have large multi-stage architectures. In contrast, we utilize simple yet efficient models for visual sound separation using only a single video frame. The results in Table~\ref{table:comparison} indicate that the model with appearance attention module can greatly improve the sound separation performance, and the systems with the appearance classifier and the category embeddings outperform recent approaches by a large margin.

\section{Conclusion}

In this paper, we studied simple yet efficient models for visual sound separation using only a single video frame and we investigated how the category information of the sound sources can be exploited in sound source separation. We experimented two types of configurations: the category labels are available at the training time; or we know if the training sample pairs are from same or different category. The extensive evaluations demonstrated that the proposed appearance attention module efficiently enhances the distinctions of the predicted source type embeddings; the models with the appearance classifier and the category embeddings surpass recent baseline models and suggest how much the existing appearance network could be further improved.


\bibliographystyle{IEEEbib}
\bibliography{ms}

\end{document}